# AN EFFICIENT AUTOMATIC MASS CLASSIFICATION METHOD IN DIGITIZED MAMMOGRAMS USING ARTIFICIAL NEURAL NETWORK

Mohammed J. Islam[1], Majid Ahmadi[2] and Maher A. Sid-Ahmed[3]

{islam1l, ahmadi, ahmed}@uwindsor.ca

Department of Electrical and Computer Engineering, University of Windsor,
401 Sunset Avenue, Windsor, ON N9B 3P4, Canada

## ABSTRACT

*In this paper we present an efficient computer aided mass classification method in digitized mammograms using Artificial Neural Network (ANN), which performs benign-malignant classification on region of interest (ROI) that contains mass. One of the major mammographic characteristics for mass classification is texture. ANN exploits this important factor to classify the mass into benign or malignant. The statistical textural features used in characterizing the masses are mean, standard deviation, entropy, skewness, kurtosis and uniformity. The main aim of the method is to increase the effectiveness and efficiency of the classification process in an objective manner to reduce the numbers of false-positive of malignancies. Three layers artificial neural network (ANN) with seven features was proposed for classifying the marked regions into benign and malignant and 90.91% sensitivity and 83.87% specificity is achieved that is very much promising compare to the radiologist's sensitivity 75%.*

## 1. INTRODUCTION

Breast cancer continues to be a public health problem in the world specifically in western and developed countries. In the European Union and the United States, it is the leading cause of death for women in their 40's [1] and second in Canada after lung cancer. In 2009, an estimated 22,700 Canadian women and 170 men were diagnosed with breast cancer and 5,400 women and 50 men died from it. Therefore, 1 in 9 women (11%) is expected to develop breast cancer during her lifetime (by age 90) and 1 in 28 will die from it. Although the breast cancer occurrence rates have increased over the years, breast cancer mortality has declined among women of all ages [1]. This positive trend in mortality reduction may be associated with improvements made in early detection of breast cancer, treatment at an earlier stage and the broad adoption of x-ray mammography [1]. However, there still remains significant room for improvements to be made in x-ray mammography since they are predominantly based on the ability of expert radiologists in detecting abnormalities.

Mammography has been one of the most reliable methods for early detection of breast carcinomas. X-ray mammography is currently considered as standard procedure for breast cancer diagnosis. However, retrospective studies have shown that radiologists can miss the detection of a significant proportion of abnormalities in addition to having high rates of false positives. The estimated sensitivity of radiologists in breast cancer screening is only about 75% [2]. Double reading has been suggested to be an effective approach to improve the sensitivity.

But it becomes costly because it requires twice as many radiologists' reading time. This cost will be quite problematic considering the ongoing efforts to reduce costs of the health care system. Cost effectiveness is one of the major requirements for a mass screening program to be successful. The ultimate diagnosis of all types of breast disease depends on a biopsy. In most cases the decision for a biopsy is based on mammography findings. Biopsy results indicate that 65-90% of suspected cancer detected by mammography turned out to be benign [3]. Therefore, it would be valuable to develop a computer aided method for mass classification based on extracted features from the region of interests (ROI) in mammograms. This would reduce the number of unnecessary biopsies in patients with benign disease and thus avoid patients' physical and mental suffering, with an added bonus of reducing healthcare costs. The principal stages of computer-aided breast cancer detection and classification is shown in figure 1. After pre-processing the mammogram for example x-ray level removal, pectoral muscle removal and mass segmentation using histogram analysis, the ROIs extracted from the breast region are shown in figure 2 and masses extracted from the ROIs both for malignant and benign are shown in figure 3 that need to be classified and it is the main objectives of the paper.

In this paper automatic mass classification into benign and malignant is presented based on the statistical and textural features extracted from mass from the breast region using proposed ANN. This paper is organized as follows. Section 2 briefly reviews some existing techniques for mass classification followed by artificial neural network (ANN) in section 3. Statistical texture features are described in section 4. Section 5 describes the materials and proposed methods for mass classification. Section 6 demonstrates some simulation results and their performance evaluation finally conclusions are presented in section 7.

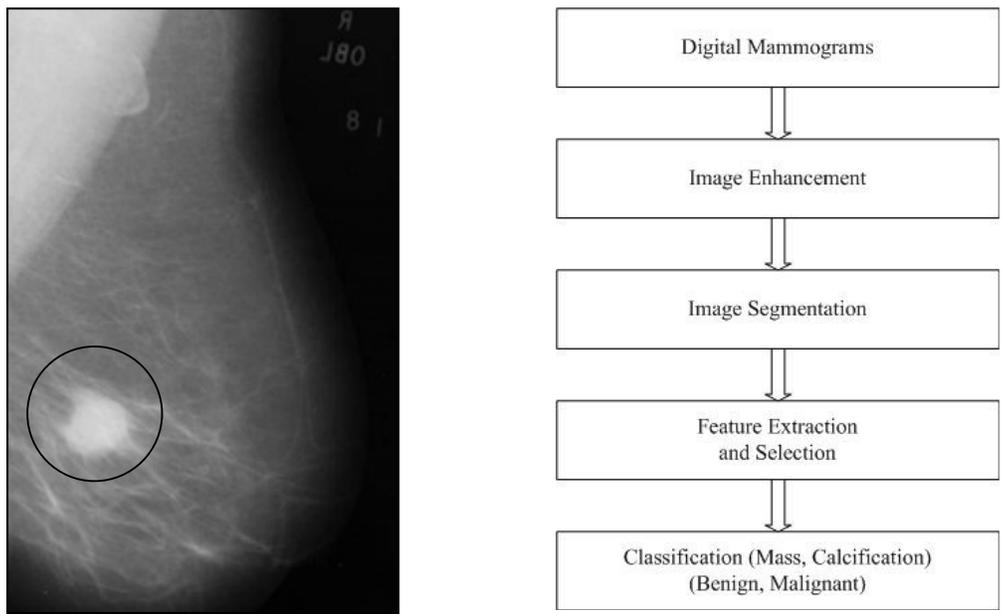

Figure 1: Sample mammogram and principal stages of breast cancer detection

## 2. LITERATURE SURVEY

Masses are one of the important early signs of breast cancer. They are often indistinguishable from the surrounding parenchyma because their features can be obscured or similar to the normal inhomogeneous breast tissues. The gray levels of those inhomogeneous tissues in the breast could vary with the distribution of breast soft tissue. Furthermore, the difficulty could be increased due to the fact that the masses in digitized mammograms are similar to the glands, cysts or dense portion of the breast [4]. This makes the automatic mass detection, segmentation and classification challenging. The main aim of this paper is to develop a classifier for breast cancer detection of masses in mammograms.

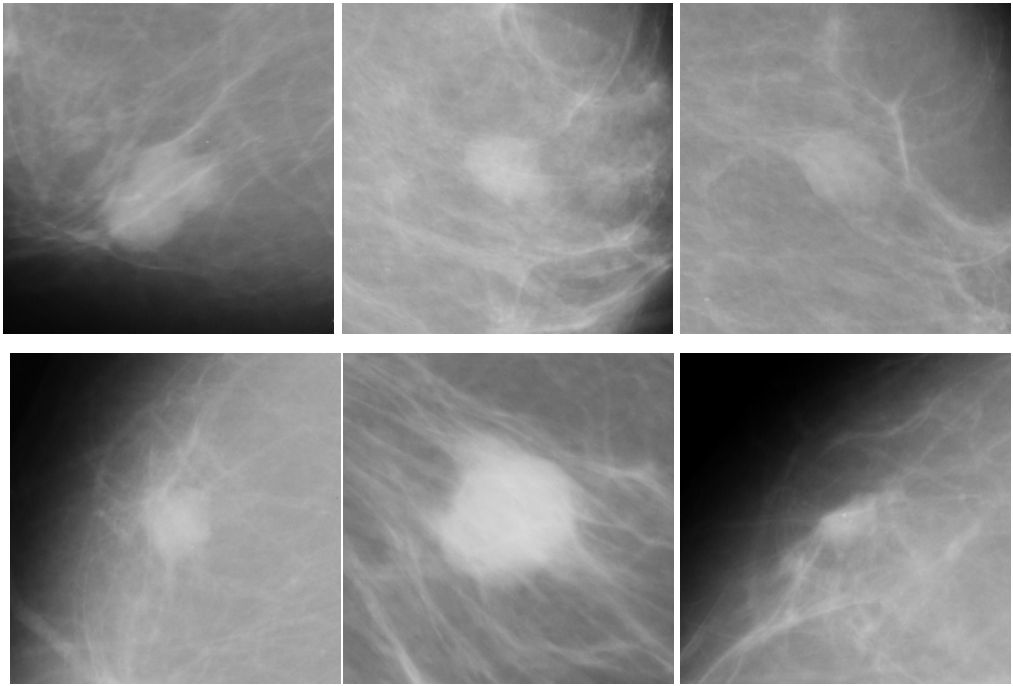

Figure 2: Sample ROIs extracted from breast region

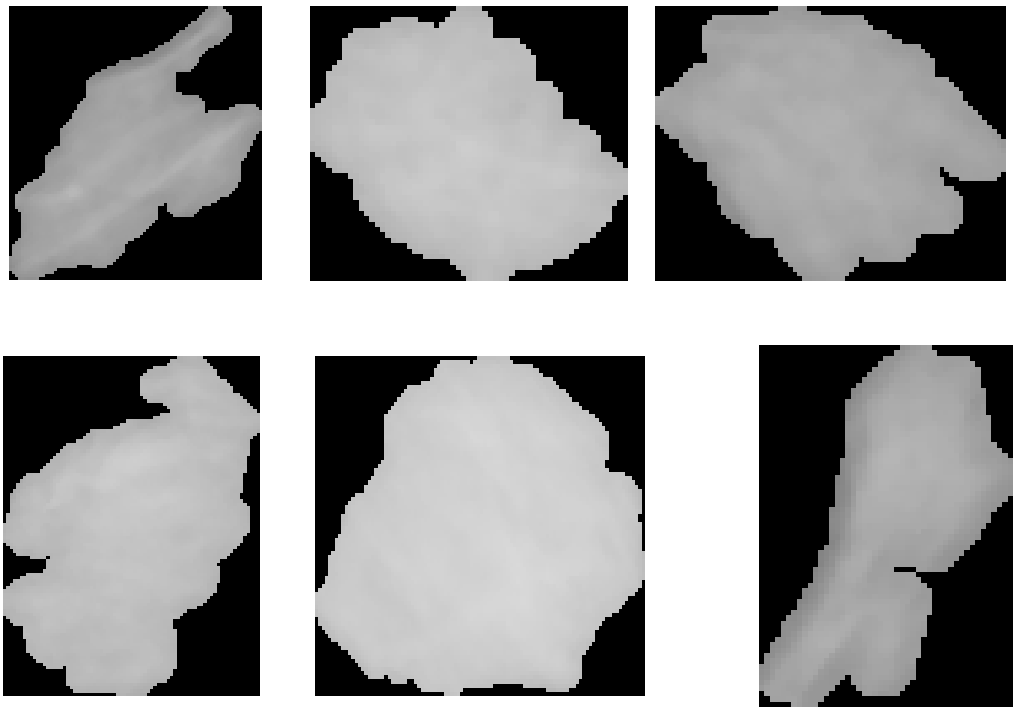

Figure 3: Sample masses extracted from the ROIs

In recent years, a few researchers in either academia or industry have used different approaches to do the classification of masses. The classification step is vital for the performance of the computer aided diagnosis (CAD) system that is shown in figure 1. Many involve the use of classification techniques to classify the marked region in the mammogram. Classifiers such as linear discriminants analysis (LDA) and ANN have performed well in mass classification. Other classifiers Bayesian network and binary decision tree and support vector machine (SVM) are also used in this application.

LDA's [3, 5, 6] are traditional method for classification. They construct decision boundaries by optimizing certain criteria to classify cases into one of mutually exclusive classes. They show high performance for linear separable problems but poor for non linear separable data. LDA in combination with stepwise feature selection [3, 7] was trained and tested on morphological features extracted using the machine segmentation and radiologist segmentation and $A_z$ area under the receiver operating characteristics (ROC) 0.89 was obtained whereas for speculation features it was 0.88.

Bayesian network uses a probabilistic approach to determine the class conditional probability density functions for background and tumor in breast cancer detection application. In [8] two phases hierarchical scheme is used to classify the masses where Bayesian classifier exists in each level. In first phase the speculated masses are discriminated from nonspeculated masses. In the second phase masses with fuzzy edges are separated from well defined edges among the nonspeculated edges.

ANNs are invaluable tools in various medical diagnostic systems. Lisboa [9] reviewed the improvements in health care arising from the participation of NN in medical field. The key attributes like distributed representations, local operations, and non-linear processing make ANN suitable for taking few difficult decisions from massive amount of data. Thus when expert knowledge is unavailable in full-fledged sense as for example in case of masses, ANN provides alternative and better solutions. Moreover, ANN constructively makes use of trained data set to make complex decisions. It is robust, no rule or explicit expression is needed and widely applicable [3]. But there is no common rule to determine the size of the ANNs, long training time and sometimes over training.

Alginahi et. al. [10, 11] developed ANN-based technique for thresholding composite digitized documents with non-uniform and complex background. It was used in the application of segmenting bank cheques from complex background for application in OCR. This method uses 8 statistical and textural features of an image.

A common database and the same genetic algorithm were used to optimize both the Bayesian belief network and neural network in [3, 12, 13]. The results show that the performance of the two classifiers converged to the same level. Therefore, it is obvious that the performance of CAD systems mostly depends on feature selection and training database than the classifiers.

In this paper, ANN-based classifier is proposed for mass classification and statistical and texture features are used as an input to the classifier. The ANN and the statistical and texture features associated with ANN-based thresholding method are described in the sections 3 and 4 respectively.

## 3. ARTIFICIAL NEURAL NETWORK

ANN is a powerful classifier that represents complex input/ output relationships. It resembles human brain in acquiring knowledge through learning and storing knowledge within inter-neuron connection strengths. Commonly, ANN's synaptic weights are adjusted or trained so that a particular input leads to a specific desired or target output. Figure 2 shows the block diagram for a supervised learning ANN, where the network is adjusted based on comparing neural network output to the desired output until the network output matches the desired output. Once the network is trained it can be used to test new input data using the weights provided from the training session.

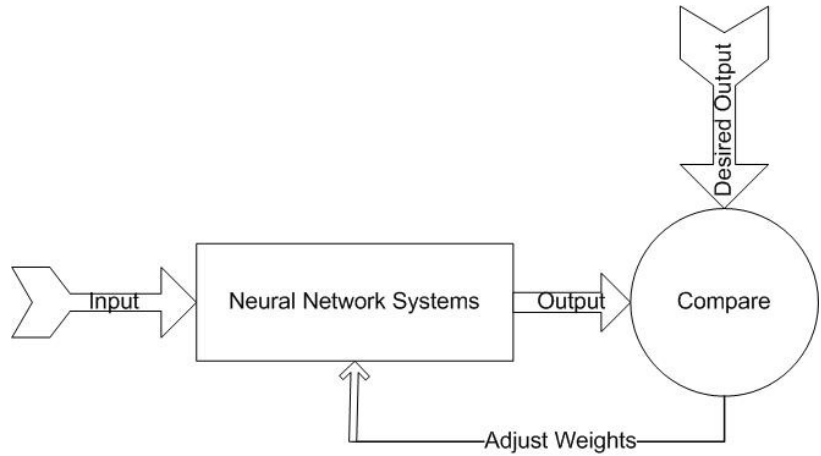

Figure 4: Supervised learning of ANN

### 3.1. Multi-layer Perceptron (MLP)

Multi-layer perceptron (MLP) is the most common neural network model. It uses supervised training methods to train the network and is structured hierarchically of several perceptrons. MLPs contain normally three layers: the input layer, hidden layers and output layer that is obvious from the figure 3. The training of such a network is complicated than single perceptron. That is why when there exists an output error; it is hard to say how much error comes from the different nodes of input, hidden and output layers and how to adjust the weights according to their contributions [14]. This problem can be solved by finding the effect of all the weights in the network by using the back-propagation algorithm [10] which is the generalization of the least-mean-square (LMS) algorithm. The input nodes, hidden nodes and the output nodes are connected via variable weights using feed-forward connections. The calculated output is compared with the target output. The total mean square error (MSE) shown in equation 1, is computed using all training patterns of the calculated and target outputs.

$$MSE = \frac{1}{2} \sum_{j=1}^{m} \sum_{i=1}^{k} \left(T_{ij} - O_{ij}\right)^2 \qquad (1)$$

Where $m$ is the number of examples in the training set, $k$ is the number of output units, $T_{ij}$ is the target output value (either 0.1 or 0.9) of the $i^{th}$ output unit for the $j^{th}$ training example, and $O_{ij}$ is the actual real-valued output of the $i^{th}$ output unit for the $j^{th}$ training example.

The back-propagation algorithm uses an iterative gradient technique to minimize the MSE between the calculated output and the target output. The training process is initialized by setting some small random weights. The training data are repeatedly presented to the neural network and weights are adjusted until the MSE is reduced to an acceptable value. A neural network has the ability to extract patterns and detect trends that are too complex to be noticed by either humans or other computer techniques.

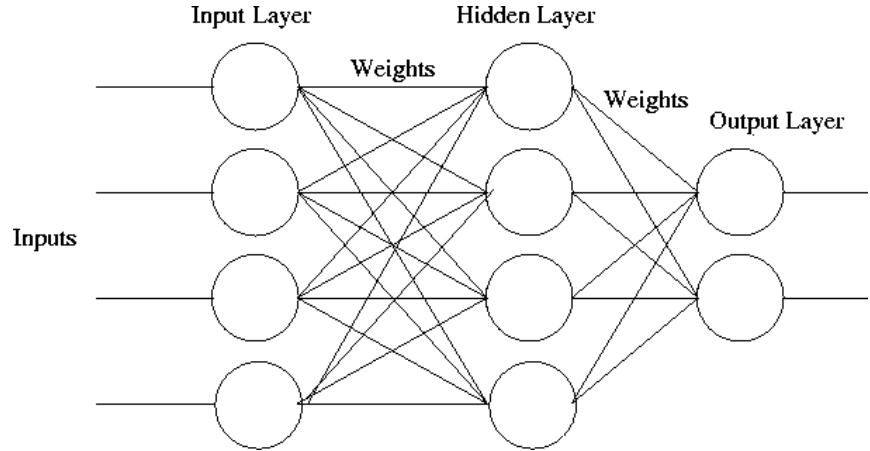

Figure 5: Multi-layer feed-forward neural network

## 4. STATISTICAL AND TEXTURE FEATURES

Statistical and texture features are extracted for each ROI. The extracted features are then used in neural network classifier to train it for the recognition of a particular ROI of similar nature. These features are pixel value, mean, standard deviation, smoothness, entropy, skewness, kurtosis and uniformity. These are adapted from [10, 11, 14, 15]. Usually these features are extracted for each centred pixel centred around the window size (2n+1)x(2n+1). In this application, same features are extracted in a modified way. In stead of extracting each pixel, the features are extracted for the whole image by treating the whole image as a pixel.

### 4.1. Mean Value

The mean, $\mu$ of the pixel values in the defined window, estimates the value in the image in which central clustering occurs.

$$\mu = \frac{1}{MN} \sum_{i=1}^{M} \sum_{j=1}^{N} p(i,j)$$

Where *p(i,j)* is the pixel value at point *(i,j)* of an image of size MxN.

### 4.2. Standard Deviation

The standard deviation, $\sigma$ is the estimate of the mean square deviation of grey pixel value *p(i,j)* from its mean value $\mu$. Standard deviation describes the dispersion within a local region. The standard deviation is defined as:

$$\sigma = \sqrt{\frac{1}{MN} \sum_{i=1}^{M} \sum_{j=1}^{N} (p(i,j) - \mu)^2}$$

### 4.3. Smoothness

Relative smoothness, *R* is a measure of grey level contrast that can be used to establish descriptors of relative smoothness.

$$R = 1 - \frac{1}{1+\sigma^2}$$

Where, $\sigma$ is the standard deviation of the image.

### 4.4. Entropy

Entropy, *h* can also be used to describe the distribution variation in a region. Overall entropy of the image can be calculated as:

$$h = -\sum_{k=0}^{L-1} Pr_k (\log_2 Pr_k)$$

Where, $Pr_k$ is the probability of the k-th grey level, which can be calculated as $Z_k / MxN$, $Z_k$ is the total number of pixels with the k-th grey level and L is the total number of grey levels.

### 4.5. Skewness

Skewness, *S* characterizes the degree of asymmetry of a pixel distribution in the specified window around its mean. Skewness is a pure number that characterizes only the shape of the distribution.

$$S = \frac{1}{MN} \sum_{i=1}^{M} \sum_{j=1}^{N} \left[ \frac{p(i,j) - \mu}{\sigma} \right]^3$$

Where, $p(i,j)$ is the pixel value at point *(i,j)*, $\mu$ and $\sigma$ are the mean and standard deviation respectively.

### 4.6. Kurtosis

Kurtosis, *K* measures the peakness or flatness of a distribution relative to a normal distribution. The conventional definition of kurtosis is:

$$K = \left\{ \frac{1}{MN} \sum_{i=1}^{M} \sum_{j=1}^{N} \left[ \frac{p(i,j) - \mu}{\sigma} \right]^4 \right\} - 3$$

Where, $p(i,j)$ is the pixel value at point *(i,j)*, $\mu$ and $\sigma$ are the mean and standard deviation respectively. The -3 term makes the value zero for a normal distribution.

### 4.7. Uniformity

Uniformity, *U* is a texture measure based on histogram and is defined as:

$$U = \sum_{k=0}^{L-1} Pr_k^2$$

Where, $Pr_k$ is the probability of the k-th grey level. Because the $Pr_k$ have values in the range [0, 1] and their sum equals 1, U is maximum in which all grey levels are equal, and decreases from there. Before computing any of the descriptive texture features above, the pixel values of the image were normalized by dividing each pixel by 255 in order to achieve computational consistency.

## 5. PROPOSED METHODS

ANN exploits the major mammographic characteristics texture to classify the mass into benign or malignant. To fulfil the objectives of this paper ANN uses mean, standard deviation, entropy, skewness, kurtosis and uniformity that is described in section 4. These 7 features are used in preparing the training data for multi-layer perceptron (MLP) neural network which are obtained from the whole extracted mass region. To do that an image file of mass (benign or malignant) is loaded and the 'Features Calc" button is pressed that is shown in figure 6. Seven features of the loaded image is calculated. Based on the prior information either benign (0) or malignant (1) button is selected and corresponding 0 or 1 is placed in the output field. The calculated 7 features and their corresponding target value (for benign=0 and malignant=1) are stored in a file. The same process is repeated for more masses both malignant and benign. This way all the training samples are stored in the file that is used as inputs to the ANN to train the network to produce the weights needed for testing the classifier.

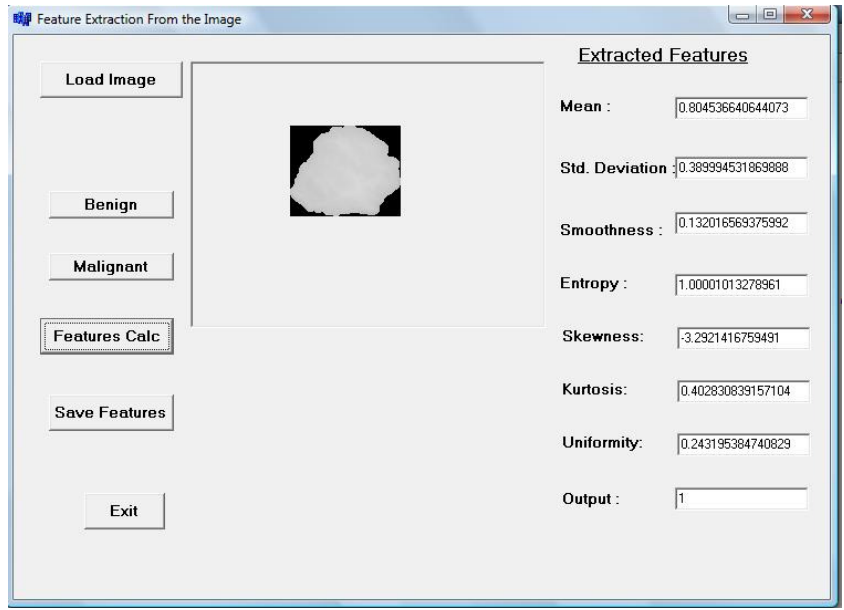

Figure 6: Training data preparation

Once the training samples are available, these samples are used to train the network. In that case, number of input units, hidden units and output unit must be entered. The proposed artificial neural network structure is shown in table 1. Since 7 features are used in that application, therefore number of input units are 7, output is either benign or malignant that can be represented by 1 unit. Hidden units are calculated using the expression shown in table 1. There is no straight forward rule to select the number of hidden units. However, the optimal number of features are (input+output)*2/3 that is also verified by the well known DTREG [17], a commercial predictive modelling software. Once the training is done, the weight vector is used to test the proposed ANN. In that case weights are kept constant. Since weights are not changing during testing it is called offline processing. The sample screen capture of testing the ANN is shown in figure 7.

Table 1: Proposed ANN structure for mass classification

| Input Units | n | 7 |
|---|---|---|
| Hidden Units | $(n+1)*\frac{2}{3}$ | 5 (Verified by DTREG[17]) |
| Output Unit | 1 | 1 |
| Weights | $n*\left((n+1)*\frac{2}{3}\right)+(n+1)*\frac{2}{3}$ | 40 |

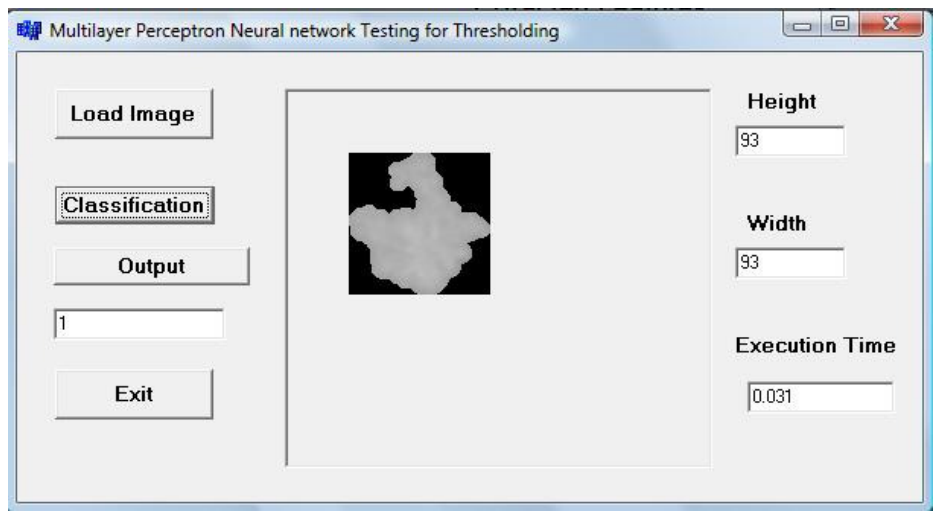

Figure 7: Testing the proposed ANN classifier

## 6. SIMULATION RESULTS AND PERFORMANCE EVALUATION

### 6.1. Image Database

To develop and evaluate the proposed system we used the Mammographic Image Analysis Society (MiniMIAS) [16] database. It is an organization of UK research group. Films were taken from UK National Breast Screening Programme that includes radiologist's "truth"-markings on the locations of any abnormalities that may be present. Images are available online at the Pilot European Image Processing Archive (PEIPA) at the University of Essex. This database contains left and right breast images for a total of 161 (322 images) patients with ages between 50 and 65. All images are digitized at a resolution of 1024x1024 pixels and at 8-bit grey scale level. The existing data in the collection consists of the location of the abnormality (like the center of a circle surrounding the tumor), its radius, breast position (left or right), type of breast tissues (fatty, fatty-glandular and dense) and tumor type if it exists (benign or malign). Each of the abnormalities has been diagnosed and confirmed by a biopsy to indicate its severity: benign or malignant. In this database, 42 images contain abnormalities (malignant masses) and 106 images are classed as normal and rest of them either contains microcalcifications or benign.

## 6.2. Results and Performance

The table 1 shows the proposed ANN structure. The proposed structure has 3 layers comprised of 7 units in input layer, 5 units in hidden layer and 1 unit in the output layer. 5 units in hidden layer are optimal number of hidden layers that is verified by the DTREG commercial software [17]. Therefore, the total weights become 40. Total 69 correctly segmented masses are used for classification where 25% images are used for training and 75% are used for testing purpose and the overall classification for benign is 83.87% and for malignant 90.91% that is shown in table 2.

Three examples are shown as an output of the proposed ANN classifier in figure 8, 9 and 10. In figure 8 and 9, masses are correctly classified as benign and malignant respectively whereas in figure 10 original mass is malignant but it is classified as benign using the proposed method.

Table 2: Simulation Results: Mass Classification, Sample Size: 69

| Correct Classification (%) | | Misclassification (%) | | Radiologist Misclassification (%) | |
|---|---|---|---|---|---|
| Benign | Malignant | Benign | Malignant | Benign | Malignant |
| 83.87 | 90.91 | 16.63 | 9.09 | 65-90 | Not available |

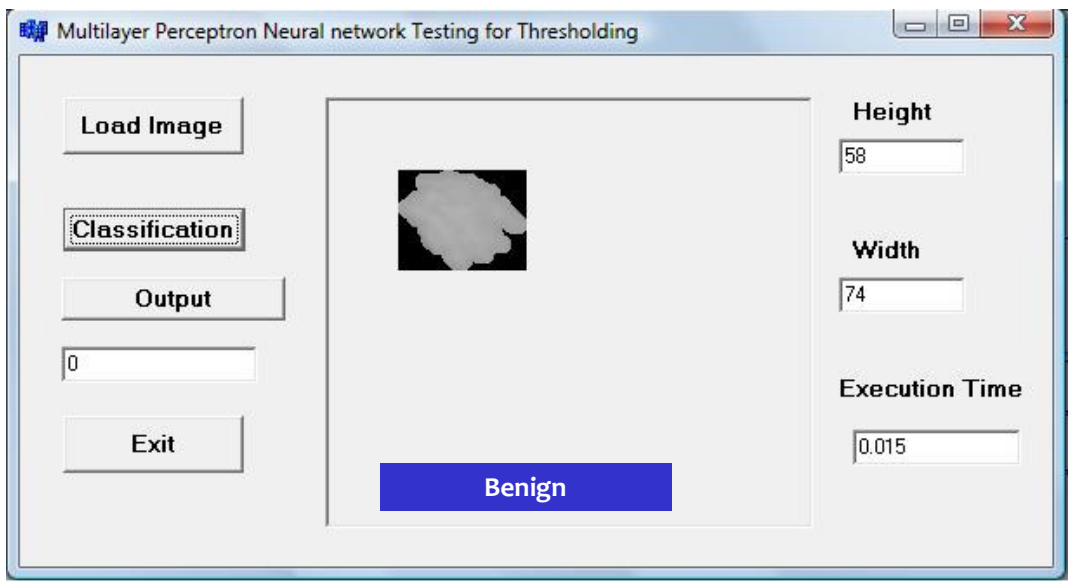

Figure 8: Mass classification: Example 1: Correct classification (Benign mass)

The classification performance can be assessed in terms of the sensitivity and specificity of the system. Sensitivity (SN) is the proportion of actual positives which are correctly identified and it is mathematically defined in equation 2 and specificity (SP) is the proportion of negatives which are correctly identified and is mathematically defined in equation 3. The sensitivity and specificity of the proposed system compared to the radiologist's sensitivity is shown in table 3.

$$SN = \frac{TP}{TP + FN} \quad (2)$$

$$SP = \frac{TN}{TN + FP} \quad (3)$$

Where, TP- True positive, TN- true negative, FP-false positive and FN- false negative.

Table 3: Comparison of the sensitivity of the proposed system to the radiologist sensitivity

| TP | TN | FP | FN | SN (%) | SP (%) | Radiologist Sensitivity (%) |
|----|----|----|----|--------|--------|------------------------------|
| 20 | 26 | 5  | 2  | 90.91  | 83.87  | 75                           |

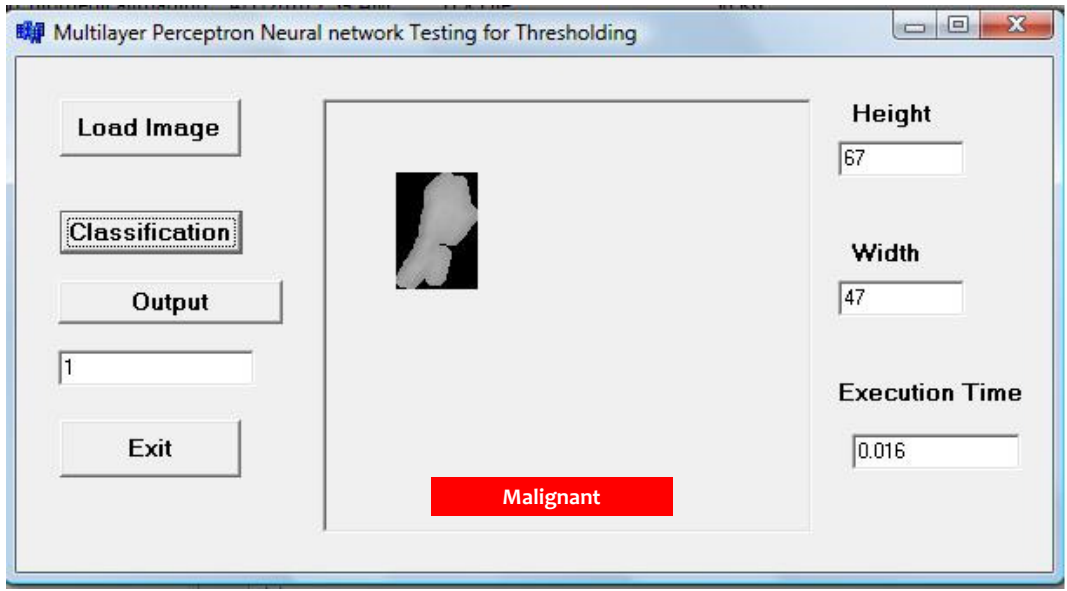

Figure 9: Mass classification: Example 2: Correct classification (Malignant mass)

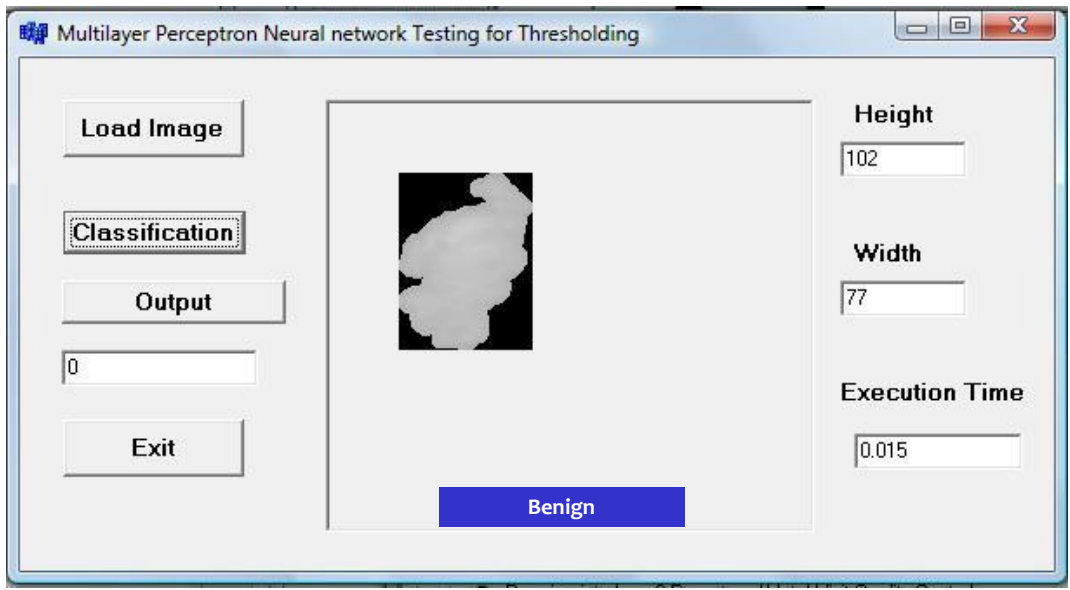

Figure 10: Mass classification: Example 3: Misclassification (classified as benign of a Malignant mass)

## 7. CONCLUSIONS

Mass classification is a vital stage for the performance of the computer aided breast cancer detection. It reduces the false positive rate by reducing the unnecessary biopsy and health care cost as well. Different classifiers were used in biomedical imaging application like breast cancer detection from mammograms. However, ANN shows very good performance in medical diagnostic systems. In this paper, an artificial neural network of 3 layers is proposed for mass classification. The performance of the proposed structure is evaluated in terms of efficiency, adaptability and robustness. Computational time is around 15~20 ms for each mass classification. It was evaluated on 69 images containing malignant and benign masses with different size, shape and contrast. The algorithm works properly in all cases and the proposed structure was evaluated with and without preliminary denoising steps. In both cases results are found to be comparable. Using the proposed ANN classifier, 90.91% sensitivity and 83.87% specificity is achieved which is very much encouraging compare to the radiologist's sensitivity 75%.

## ACKNOWLEDGEMENT

This research has been financially supported by Ontario Student Assistance Program through Ontario Graduate Scholarship (OGS), Casino Windsor Cares/Gail Rosenblum Memorial Breast Cancer Research Scholarships and Research Centre for Integrated Microsystems (RCIM) of University of Windsor. The financial support of these organizations is greatly acknowledged with appreciation.

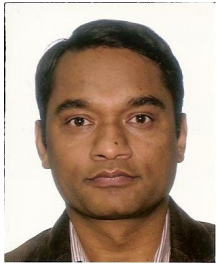

**Mohammed J. Islam** was born in Comilla, Bangladesh on June 1, 1975. He received the BSc. (Hon's) and MSc. in Electronics and Computer Science in 1995 and 1996 respectively from Shahjalal University of Science and Technology, Sylhet, Bangladesh. He received MASc in Electrical and Computer Engineering from Ryerson University, Toronto, ON, Canada, in 2003. Currently he is doing PhD in Electrical and Computer Engineering at University of Windsor, Windsor, ON, Canada. The author is the IEEE student member and the member of International Association of Engineers (IAENG) and member of Professional Engineers Ontario (PEO). His research interests include but not limited to machine learning and computer vision, image processing and document image analysis.

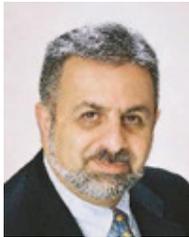

**Majid Ahmadi** received the B.Sc. degree in electrical engineering from Arya Mehr University, Tehran, Iran, in 1971 and the Ph.D. degree in electrical engineering from Imperial College of London University, London, UK, in 1977. He has been with the Department of Electrical and Computer Engineering, University of Windsor, Windsor, Ontario, Canada, since 1980, currently as University Professor and Director of Research Center for Integrated Microsystems. His research interests include digital signal processing, machine vision, pattern recognition, neural network architectures, applications, and VLSI implementation, computer arithmetic, and MEMS. He has co-authored the book Digital Filtering in 1 and 2 dimensions; Design and Applications (New York: Plennum, 1989) and has published over 400 articles in these areas. He is the Regional Editor for the Journal of Circuits, Systems and Computers and Associate Editor for the Journal of Pattern Recognition, and the International; Journal of Computers and Electrical Engineering.

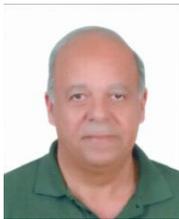

**Maher A. Sid-Ahmed** was born in Alexandria, Egypt, in March 1945. He received the BASc. degree in Electrical Engineering from the University of Alexandria, Egypt, in 1968 and the Ph.D. degree in Electrical Engineering from the University of Windsor, Windsor, Ontario, Canada, in 1974. After graduation, he worked at the Alberta Government telephones and the University of Alexandria, Egypt. He joined University of Windsor in 1978 as a faculty member and is presently the department head for the ECE program. Dr. Sid-Ahmed holds four US patents, published a book on Image processing with McGraw-Hill and authored and co-authored over 100 papers. His areas of interest include image processing, robotic vision, pattern recognition and Improved Definition Television